\documentclass[a4paper]{article}
\usepackage{times}
\usepackage{spconf}
\usepackage{graphicx}
\usepackage{amsfonts}
\usepackage{amssymb}
\usepackage{amsmath}
\usepackage{floatrow}
\usepackage{bm}
\usepackage{graphicx}
\usepackage{subfigure}
\usepackage{bbm}
\usepackage{color}

\title{Selective Attention Encoders by Syntactic Graph Convolutional Networks for Document Summarization
}

\name{ 
    Haiyang Xu$^1$, 
    Yun Wang$^3$,
    Kun Han$^2$,
    Baochang Ma$^1$,
    Junwen Chen$^1$,
    Xiangang Li$^1$
}

\address{
  $^1$DiDi Chuxing, Beijing, China\\
  $^2$DiDi Research America, Mountain View, CA, USA\\
  $^3$Peking University, Beijing, China\\
  xuhaiyangsnow@didiglobal.com,
wangyunazx@pku.edu.cn,\\
\{kunhan, maobaochang, chenjunwen, lixiangang\}@didiglobal.com
  }

\begin{document}
\maketitle

\begin{abstract}
Abstractive text summarization is a challenging task, and one need to design a mechanism to effectively extract salient information from the source text and then generate a summary. A parsing process of the source text contains critical syntactic or semantic structures, which is useful to generate more accurate summary. However, modeling a parsing tree for text summarization is not trivial due to its non-linear structure and it is harder to deal with a document that includes multiple sentences and their parsing trees. In this paper, we propose to use a graph to connect the parsing trees from the sentences in a document and utilize the stacked graph convolutional networks (GCNs) to learn the syntactic representation for a document. The selective attention mechanism is used to extract salient information in semantic and structural aspect and generate an abstractive summary. We evaluate our approach on the CNN/Daily Mail text summarization dataset. The experimental results show that the proposed GCNs based selective attention approach outperforms the baselines and achieves the state-of-the-art performance on the dataset.
\end{abstract}

\begin{keywords} 
Document Summarization, Sequence-to-Sequence, Dependency Parsing Tree, Graph Convolutional Networks
\end{keywords}

\section{Introduction}

Document summarization aims at generating a short and fluent summary consisting of the salient
information of the source text. Existing approaches for text summarization are divided into two major types: extractive and abstractive. Extractive methods produce summaries by extracting important sentences from the original document. Abstractive methods produce the summaries by summarizing the salient information of the source text and representing by arbitrary words.

The end-to-end neural framework based on sequence-to-sequence (Seq2Seq) models \cite{sutskever2014sequence} have achieved tremendous success in many text generation tasks such as machine translation \cite{luong2015effective}, dialogue systems \cite{bordes2016learning}. The essence of Seq2Seq method is an encoder-decoder framework, which first encodes the input sentence to a low dimensional representation and then decodes the abstract representation based on attention mechanism\cite{gu2016incorporating}.

Some researchers also apply neural Seq2Seq model to abstractive text summarization \cite{see2017get}. Although it is straightforward to adopt the Seq2Seq approach to text summarization, there is a significant difference between it and machine translation: the important component of text summarization is capturing the salient information of the original document for generating summary instead of aligning between the input sentence and the summary. 
Explicit information selection in text summary have proven to be more effective than implicit learning only via the Seq2Seq approach. 
Recent researchers has explored selective gate based on global document information\cite{Zhou2017Selective}\cite{li2018improving}, and combining extractive method \cite{tan2017abstractive}\cite{hsu2018unified} to capture salient information explicitly for decoding and achieved the current state-of-the-art results. However, these models ignore the syntactic structure of the source text which can help choose important words in structure to generate more accurate summary, or feature-based models \cite{song2018structure} represent dependency information by hand-crafted features, which face the problem of sparse feature spaces and not emphasizing the explicit importance of structure.

In this paper, we explore syntactic graph convolutional
networks (GCNs)\cite{marcheggiani2017encoding} to model non-euclidean document structure 
and adopt attention information gate to select salient information for generating text summarization. 
Specifically, We build a document-level graph with heterogeneous types of nodes
and edges, which are formed by connecting the dependency parsing trees from the sentences in a document. We adopt stacked convolutional neural networks (GCNs) to learn the local and non-local syntactic representation for a document, which have proved the effectiveness in other NLP tasks\cite{marcheggiani2017encoding}\cite{zhang2018graph}\cite{bastings2017graph}. Then, we employ attention mechanism to acquire global document representation combining syntactic and semantic information, and use explicit information selection gate based on global document representation to choose important words for generating better summary. We evaluate our model to the CNN/Daily Mail text summarization datasets. The experimental results show that the proposed GCNs encoders model outperforms the state-of-the-art baseline models.

\section{Related Work}
Existing approaches for document summarization are divided into two major categories: Extractive and Abstractive.

\textbf{Extractive:} \cite{nallapati2017summarunner} use hierarchical Recurrent Neural Networks (RNNs) to get the representations of the sentences and classify the importance of sentences. \cite{narayan2018ranking} rank extracted sentences for summary generation through a reinforcement learning and \cite{chen2018fast} extract salient sentences and propose a new policy gradient method to rewrite these sentences (i.e., compresses and paraphrases) to generate a concise overall summary. \cite{narayan2018document} propose a framework composed of a hierarchical document encoder based on CNNs and an attention-based extractor with attention over external information.\cite{zhou2018neural} present a new extractive framework by joint learning to score and selecting sentences.

\textbf{Abstractive:}\cite{rush2015neural} firstly apply neural networks for text summarization by using a local attention-based model to generate word conditioned on the input sentence.
\cite{nallapati2016abstractive} applies Seq2Seq framework with hierarchical attention for text summarization. \cite{tan2017abstractive} proposes graph-based attention mechanism to summarize the salient information of document. 
However the above neural models all faces out-of-vocabulary (OOV) problems since the vocabulary is fixed at training stage. In order to solve this problem, Point Network \cite{vinyals2015pointer}, \cite{see2017get} and CopyNet \cite{gu2016incorporating} have been proposed to allow both copying words from the original text and generating arbitrary words from a fixed vocabulary. \cite{hsu2018unified} propose a unified model via inconsistency loss to combine the extractive and abstractive methods. \cite{gehrmann2018bottom} adopt bottom-up attention to alleviate the issue of Point Network tending to copy long-sequences.
Recently, more and more researchers\cite{Zhou2017Selective}\cite{li2018improving} focus on explicit information selection in encoding step, which filters unnecessary words and uses important words for generating summary. The focus of our model is selecting salient syntactic and semantic information.

\begin{figure}[h]
 	\centering
	\includegraphics[height=0.9\linewidth,width=1.0\linewidth]{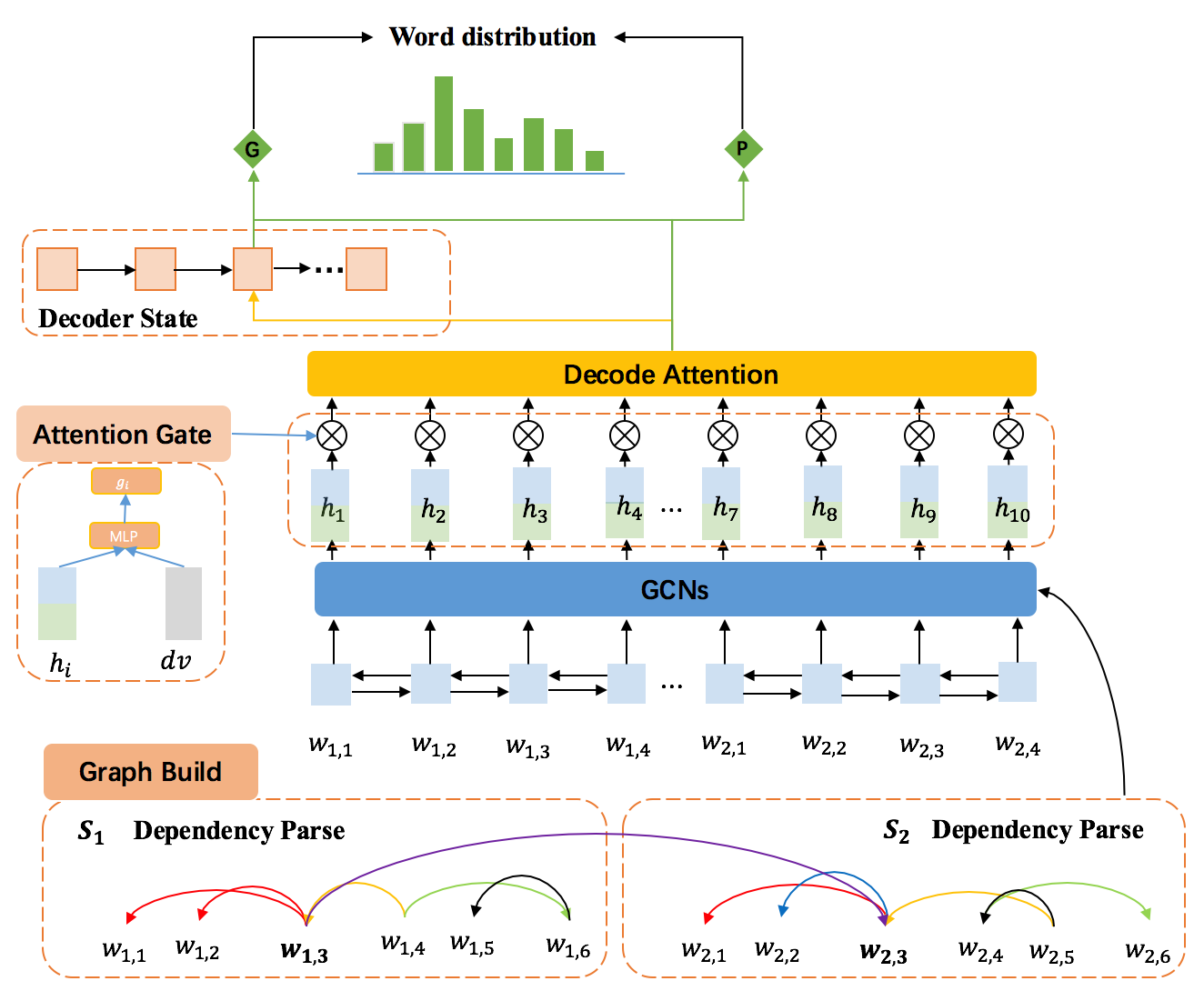}
	\caption{Overall architecture of the proposed model. The structural document graph(different colors denote the types of nodes and edges) represented by GCNs and the semantic document representation represented by BiLSTM are combined via attention mechanism to get document information. Then, selective gate filters unnecessary words for generating the summary.}
	\label{fig:model}
\end{figure}
\section{Algorithm Details}
The architecture of our model is shown in Figure~\ref{fig:model}. In this section, we describe the proposed model specifically, which consists of Semantic Document Encoder, Syntactic Document Encoder with GCN, and Attention Information Gate.

\subsection{Semantic and Syntactic Document Encoder with GCN}
\subsubsection{Semantic Document Encoder}
Given a document $d = <w_1, w_2, ..., w_n>$ concatenating all sentences into a long sequence, where $w_{i}$ is $i_{th}$ words in the document and $n$ is the sequence length of document, 
we employ a bidirectional long short-term memory (BiLSTM) \cite{graves2005framewise} as the encoder. The BiLSTM consists of forward LSTM $\overrightarrow{f}$, which reads the document $d$ from $w_{1}$ to $w_{n}$ and backward LSTM $\overleftarrow{f}$ reads the document $d$ from $w_{n}$ to $w_{1}$:
\begin{align}
&x_{i} =  W_{e}w_{i},i\in\{1,..,n\}\\
&\overrightarrow{h_{i}^{e}}=\overrightarrow{LSTM}(x_{i}), i\in\{1,..,n\}\\  
&\overleftarrow{h_{i}^{e}}=\overleftarrow{LSTM}(x_{i}),i\in\{1,..,n\}
\end{align}
Where $x_{i}$ is the distributed representation of token $e_{i}$ by embedding matrix $W_{e}$. We concatenate every forward hidden state $\overrightarrow{h_{j}^{e}}$ with the backward hidden state $\overleftarrow{h_{j}^{e}}$ to get the original word semantic representation $h_{i}^{e}=[\overrightarrow{h_{i}^{e}},\overleftarrow{h_{i}^{e}}]$. 


\subsubsection{Syntactic Document Encoder}
In order to build a document-level graph, we apply a parser to generate the dependency tree $l_k$ of every sentence $s_{k}$ by treating each syntactic dependency direction and label as a different edge type. Directions and labels of edges can discriminate the important words in structure and our comparing experimental results also demonstrate the essence of learning directions and labels of edges in text summarization task.  Then, we link the syntactic root node serially with adjacent sentence type
edges to build a complete document adjacency matrix.

We apply GCNs \cite{marcheggiani2017encoding} \cite{bastings2017graph}to compute the structural representation of every word on the constructed document graph, which keeps separate parameters
for each edge type. Furthermore, we stack many GCNs layers to make every node conscious of more distant neighbors. After $L$ layers, every node representation can capture local and global syntactic information. 

Specifically, in the $L-$layer GCN, we denote $h^{s}_{l-1,i}$ , $h^{s}_{l,i}$ as the $i_{th}$ input and output vector of node $i$ at the $l_{th}$ layer, the $1_{th}$ input vector is the word semantic representation $h_{i}^{e}$.
A graph convolution operation can be denoted as follows:
\begin{align}
h^{s}_{l,i} =  &\sigma(\sum_{j \in M(i)}
W^{(l)}_{(i,j)}h^{s}_{l-1,i}  + b^{(l)}_{(i,j)})
\end{align}
Where $W^{(l)}_{(i,j)}$, $b^{(l)}_{(i,j)}$ are the trainable parameters, $M_{i}$ is the set of neighbouring nodes of $i_{th}$ node. As \cite{marcheggiani2017encoding}, we set the word semantic representation $h_{i}^{e}$ as the original inputs of GCNs $h_{0,i}^{s}$ to alleviate the parsing error.  Furthermore, we also prevent over-parametrization by setting the same weighs of direction and label edges, but separate weights for adjacent sentence edges and sole bias for all edge types. 

We use the output vector of node $i$ at the $l_{th}$ layer as word structural representation $h_{i}^{s}=h_{l,i}^{s}$. We concatenate the semantic and structural representation to get the informative word representation $h_{i} = [h_{i}^{e}, h_{i}^{s}]$.


  
\subsection{Attention Information Gate}
We propose a novel global information gate based on attention mechanism to select the salient information from the input.
Concretely, we adopt the attention mechanism\cite{yang2016hierarchical} aggregate the representation of those
informative words to form a document vector. 

Then, the gate network takes the document vector $dv$ and the word representation as the input to compute the selective gate vector $g_{i}$:
\begin{align}
&u_{i} = tanh(W_{w}h_{i}+b_{w})\\
&a_{i} = \frac{exp(u_{i}^{T}u_{w})}{\sum_{n}exp(u_{i}^{T}u_{w})}\\
&dv = \sum_{n}a_{i}h_{i}\\
&g_{i} = \sigma(W_{g}h_{i}+U_{g}dv+b_{g}) 
\end{align}
Where $W_{w}$, $W_{g}$, $U_{g}$ are the trainable weights and $b_{w}$, $b_{g}$ are the trainable bias. Then, each word can be filtered by the gate vector $g_{i}$ to get important words for decoding:
\begin{align}
h_{i}^{*} = h_{i}  \odot g_{i}
\end{align}
Where $h_{i}^{*}$ is the representation of word $w_{i}$ after information filtration and used as the input word representation
for the decoder to generate the summary. $\odot$ is element-wise multiplication. 

In the decoding stage, we apply pointer-generator network to alleviate the OOV problems and coverage network to prevent repetition\cite{see2017get} as prior works. Furthermore, we also employ bottom-up attention\cite{gehrmann2018bottom} to relieve the problem of copying very long sequences using pointer-generator network. The final loss consists of negative log-likelihood and the coverage loss.

\section{Evaluations}
\label{sec:exp}
In this section, we introduce the expermental setup and present the experimental results. 
\subsection{Experiment Setup}
We use CNN/Daily Mail dataset \cite{see2017get}~\footnote{www.github.com/abisee/pointer-generator} to evaluate our model, which consist of long-text and has been widely used in text summarization task. We used scripts supplied by \cite{nallapati2016abstractive} ~\footnote{https://github.com/abisee/cnn-dailymailr}, to produce the non-anonymized version of the CNN/Daily Mail summarization dataset, which contains 287,277 training pairs, 13,368 validation pairs and 11,490 test pairs. For all experiments, we use 50k words of the source vocabulary. To obtain the syntactic information of the sentences in the corpora, we use Stanford Parser \cite{chen2014fast}~\footnote{https://nlp.stanford.edu/software/lex-parser.shtml} to get dependency trees with 39 edge labels.

Our model takes 256-dimensional hidden states and use 200-dimensional word embeddings. We choose Adagrad (This was found to work best of Stochastic Gradient Descent, Adadelta, Momentum, Adam and RMSProp) with learning rate 0.15 and initialize the accumulator value with 0.1.
For hyper-parameter configurations, we adjust them according to the performance on the validation set and 200 examples randomly sampled from validation set in inference stage for bottom-up attention\cite{gehrmann2018bottom}~\footnote{https://github.com/sebastianGehrmann/bottom-up-summary}. After tuning, attention mask threshold is set to 0.1, the weight of length penalty is 0.4, and coverage loss weighted is $\lambda = 1$. We use 4 layers GCNs in encoder and set beam size to 4. 
\begin{table} 
\caption{Results of abstractive summarizers on the CNN-DM dataset. The first section shows extractive baselines. The second section describes abstractive approaches.
The third section presents our model. All our ROUGE scores have a 95\% confidence interval with at most 0.25.}
\centering
       \begin{tabular}{l c c c}
       \hline \hline
    Methods & R-1 & R-2 & R-L  \\
    \hline 
    \textbf{{Extractive}}  & & \\
      Lead-3 \cite{see2017get} & 40.34 & 17.70& 36.57\\
     RuNNer \cite{nallapati2017summarunner} & 39.60 & 16.20 &35.30 \\ 
     Refresh \cite{narayan2018document} & 40.00 & 18.20& 36.60 \\
     RNN-RL \cite{chen2018fast} & 41.47 & 18.72& 37.76\\
     NeuSUM \cite{zhou2018neural}&41.59&19.01&37.98\\
    \hline 
    \textbf{{Abstractive}}  & & &\\
      Coverage\cite{see2017get} &39.53  & 17.28&36.38\\ 
      Intra-attn\cite{paulus2017deep} &39.87  & 15.82&36.90\\  
      Inconsistency\cite{hsu2018unified} &40.68  & 17.97&37.13\\
      Bottom-up\cite{gehrmann2018bottom} &41.22  & 18.68&38.34\\
      Info-select\cite{li2018improving} &41.54  & 18.18&36.47\\
    \hline 
    \textbf{Our model} & \textbf{41.79} & \textbf{19.06}&\textbf{38.56} \\
    \hline \hline
    \end{tabular}
    \label{tab:comp}
\end{table}.
\subsection{Experimental Results}
We adopt the widely used ROUGE~\cite{lin2004rouge} by pyrouge  ~\footnote{pypi.python.org/pypi/pyrouge/0.1.3}for evaluation metric. 
It measures the similarity of the output summary and the standard reference by computing overlapping n-gram, such as unigram, bigram and longest common subsequence (LCS). In the following experiments, we adopt ROUGE-1 (unigram), ROUGE-2 (bigram) and ROUGE-L (longest common subsequence) for evaluation.

It can be observed from Table \ref{tab:comp} that the proposed approach achieves the best performance on the CNN/Daily datasets over state-of-the-art extractive as well as abstractive baselines. Comparing with the same architecture of Coverage, our model has significant improvement on (2.26 on ROUGE-1, 1.78 on ROUGE-2, 2.18 on ROUGE-L), which demonstrates the effectiveness of our model combining the syntactic and semantic information by information gate and modeling the heterogeneous document-level graph via stacked GCNs.  

\begin{table}
\caption{Results of removing different components of our model on the CNN dataset. All our ROUGE scores have a 95\% confidence interval with at most 0.25.}
\centering
       \begin{tabular}{l c c c}
       \hline \hline
    Methods & R-1 & R-2 & R-L  \\
    \hline 
     \textbf{Our model} &\textbf{32.39}&\textbf{13.00}&\textbf{29.51}\\
     -Attn-Gate &31.91  & 12.68 &29.03 \\ 
     -GCNs & 31.46 & 12.32& 28.60 \\
    \hline \hline
    \end{tabular}
    \label{tab:diff}
\end{table}
Furthermore, to further study the effectiveness of each component of our model, we conduct several ablation experiments on the CNN dataset~\footnote{The CNN dataset is the subset of CNN/Daily dataset and is also widely used for text summarization. We also have achieved state-of-the-art performance in this dataset}. "-Atten-Gate" denotes that we remove the Attention Gate and only use all encoded word to decode; "-GCNs" denotes that we remove the GCNs and the Attention Gate, which degrades into Coverage\cite{see2017get}. Table \ref{tab:diff} shows that each component of our model all improve the performance on the dataset evidently. GCNs model the structure of document effectively and represent it in a dense low-dimension feature. Attention Gate adopt the attention mechanism to acquire informative document representation combining syntactic and semantic information, then use selective gate to detect critical words for generating the summary.


\section{Conclusions}
In this work, we propose syntactic graph convolutional encoders based on dependency trees for abstractive text summarization, which uses graph convolutional networks (GCNs) to learn the representation of syntactic structure of the source text, then adapt the attention mechanism combining semantic and structural information to select important words for decoding. We evaluate our model to the CNN/Daily Mail text summarization datasets. The experimental results show that the proposed GCNs encoders model outperforms the state-of-the-art baseline models.

\bibliographystyle{IEEEbib}
\bibliography{./selective_tree_summary}
\end{document}